# ANFIS-based prediction of power generation for combined cycle power plant

Mary Pa, Amin Kazemi

Department of Electrical Engineering, Lakehead University

Department of mechanical and industrial engineering university of Toronto

Email: mpaparim@lakeheadu.ca, amin.kazemi@utoronto.ca

**Abstract**

This paper presents the application of an adaptive neuro-fuzzy inference system (ANFIS) to predict the generated electrical power in a combined cycle power plant. The ANFIS architecture is implemented in MATLAB through a code that utilizes a hybrid algorithm that combines gradient descent and the least square estimator to train the network. The Model is verified by applying it to approximate a nonlinear equation with three variables, the time series Mackey-Glass equation and the ANFIS toolbox in MATLAB. Once its validity is confirmed, ANFIS is implemented to forecast the generated electrical power by the power plant. The ANFIS has three inputs: temperature, pressure, and relative humidity. Each input is fuzzified by three Gaussian membership functions. The first-order Sugeno type defuzzification approach is utilized to evaluate a crisp output. Proposed ANFIS is cable of successfully predicting power generation with extremely high accuracy and being much faster than Toolbox, which makes it a promising tool for energy generation applications.

## I-Introduction

Soft computing techniques have attracted many researchers in the previous few decades as it has proven to be an efficient method to deal with complicated problems for which conventional analytical methods are infeasible or too expensive. Several artificial intelligence techniques have been proposed to solve many problems in different areas. These methods include artificial neural networks (ANN), neuro-fuzzy (NF), fuzzy logic and optimization algorithms such as the genetic algorithm (GA), particle swarm optimization (PSO), artificial bee colony (ABC) algorithm and so forth. Among these techniques, ANFIS has attracted much attention and has become one of the most popular artificial intelligent models which are employed in several studies such as prediction of the energy required buildings [1], control of nonlinear systems [2], time-series predictions [3], [4], estimation of wind speed [5], membrane separation [6], to mention a few.

The Adaptive Neuro-Fuzzy Inference System technique was initially introduced by Jang in 1993 [4]. ANFIS is a hybrid artificial intelligence technique that is a combination of neural networks and fuzzy logic. Therefore, provide both the merits of the learning ability of neural networks and the knowledge representation of fuzzy systems [7], [8]. Fuzzy inference systems enable us to show uncertain situations using fuzzy If-Then rules. The system can wisely interpret the results, while other approaches, such as neural networks, cannot do so [9].

ANFIS has proven to estimate functions among other neuro-fuzzy models [10]. ANFIS utilizes several training data to map the desired output through its rule-based structure. The fuzzy sets are defined by membership functions (MF) and rules. Compounding the membership functions to each input variable converts crisp values into fuzzy values. Therefore, the input variables are fuzzy in nature, whereas the output variables are crisp in nature. In fuzzy logic, the truth of any statement has a degree of correctness contrary to classical logic, which only permits conclusions that are either true or false.

There are two main types of fuzzy inference systems that are commonly used, Mamdani and Sugeno-type. The difference between these two types is the way that the outputs are determined. In the Sugeno-type inference system, the output of the network is constant (zero-order Sugeno model) or a linear combination of the input variables (first-order Sugeno model). ANFIS is usually applied to optimize the Sugeno fuzzy inference system's parameters for matching input-output data with minimal error. For this purpose, ANFIS uses a combination of the gradient descent (GD) method and the least squares estimate (LSE). This method was first proposed by Jang [4].

This report has applied ANFIS to predict the generated electricity in a combined cycle power plant. In these types of plants, efforts are put into converting as much heat as possible from the exhaust gas of the gas turbine into steam for a steam turbine. Therefore, plant efficiency is boosted because of this heat recovery. From a set of measured data in one year, it turns out that the output electrical power of the plant ($E$) is a function of the ambient temperature ($T$), ambient pressure ($P$), and relative humidity (*H*). Here, a custom code was developed in MATLAB to construct an ANFIS structure whose inputs are *T*, *P*, and *H*. By combining the gradient descent and least square estimator optimization approach, the ANFIS has found parameters for this specific application and predict the power for some unseen data.

To ensure that the code is free of errors, w took three additional steps other than predicting and checking the output power of the plant. First, the code was used to approximate a three-variable nonlinear function for which the exact solution is known. Second, compared our ANFIS performance with the solution of the chaotic time-series data of the Mackey-Glass equation [11]. Third, implemented the same power plant model in the MATLAB ANFIS toolbox and compared the results with our custom code. The results of each step are explained in detail. Some aspects of the ANFIS, such as the effect of the type of membership functions, zero and first-order Sugeno models, number of membership functions and so forth, are discussed. The remaining of paper is as follows. Section II presents the methodology. Model evaluation is discussed in section III. Next, in section V, the discussion part is written. Finally, a conclusion is drawn in section VI, which sums up the findings and contribution of the paper concisely.

## II-Methodology

The ANFIS architecture (shown in Figure I) contains 3 inputs with 3 Gaussian membership functions assigned to each input variable. To present the ANFIS architecture, 27 fuzzy IF-THEN rules based on a first-order Sugeno model are considered. Note that higher-order Sugeno fuzzy models may be employed. However, they introduce more complexity to the system without significant merit [18].

In the design of the ANFIS, consider the maximum possible number of rules that can be stated by the "**AND**" fuzzy operator. The number of rules, in this case, can be calculated by the product of the number of membership functions of all inputs. Therefore 27 rules may be introduced to the network. could also use fewer rules to boost computational speed. However, this was not necessary since the computational time for the chosen number of training data is in the order of a few minutes. Besides, by using fewer rules, the system may find capturing the relation between the inputs and the output more challenging. Methods to find the optimal number of rules may be found elsewhere [19]. Based on the 27 rules and 9 Mfs, the total number of fitting parameters is 126, including 18 premise parameters (nonlinear) of the Mfs and 108 consequent parameters (linear) of the rules.

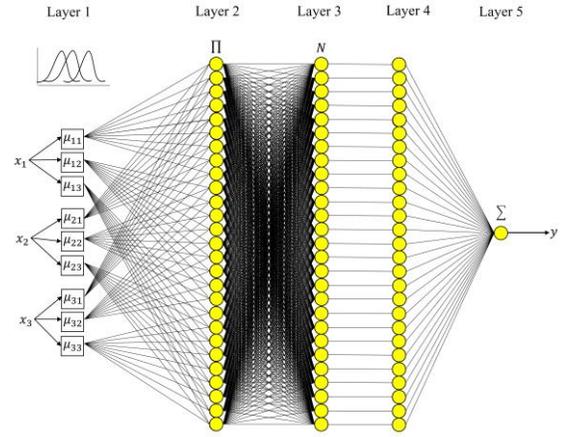

Figure-I ANFIS architecture for 3 inputs and a single output. Each input has 3 Gaussian Mfs, and 27 rules are defined

The ANFIS developed should be appropriately trained to generate an optimal input/output mapping. The iteration in each epoch consists of two significant steps. First, for the given values of the MF parameters and *P* training data pairs, the liner parameters in the consequent part are optimized using LSE. For the 27 rules considered, 108 unknowns need to be obtained. The gradient descent algorithm is employed to update the premise parameters $c_i$ and $\sigma_i$. The root mean square error (RMSE) of the model predictions is defined as:

$$E(\boldsymbol{\theta}) = \sqrt{\frac{1}{2P}\sum_{k=1}^{P}\left(y_{\text{obs}}^{(k)} - y_{\text{model}}^{(k)}\right)^2} \qquad \text{Eq. 1}$$

where $y_{\text{model}}^{(k)}$ is the predicted output for the $k^{\text{th}}$ data pair and $y_{\text{obs}}^{(k)}$ is the actual $k^{\text{th}}$ output. This function, also known as the cost function, should be minimized. The parameters of the membership functions may be updated as:

$$c_i = c_i - \eta\frac{\partial E}{\partial c_i} \qquad \text{Eq. 2}$$

$$\sigma_i = \sigma_i - \eta\frac{\partial E}{\partial \sigma_i} \qquad \text{Eq. 3}$$

where $\eta$ is the learning rate initially set to a small value (*i.e.* 0.001). It may be increased to accelerate the convergence or decreased to avoid system instability. The selection of $\eta$ is problem-specific and needs to be determined by the user. The procedure of the selection of an appropriate $\eta$ in our study will be discussed later. Initially, 80% of the data (*P* = 2521) were loaded for training the network. FIS was generated by choosing Gaussian Mfs by using a hybrid optimization algorithm. To facilitate convergence, the initial parameters of the membership functions were selected in such a way that the centers of the MFs are equally spaced along with the range of each input variable. Also, the initial membership functions meet the condition of $\epsilon$-completeness with $\epsilon = 0.5$, meaning that within the ranges of the inputs, there is always a linguistic variable for which $\mu \geq 0.5$. Once the ANFIS model training was completed, the remaining 20 % of the data was used to test the Model. The performance of the Model was verified by calculating the RMSE values during both the training and testing process.

## III-Evaluation

In this section, the custom ANFIS model is evaluated for 4 cases. First, use an analytical nonlinear equation whose exact solution is known. The Model was trained by a set of synthetic data extracted from the equation. Then apply the Model to predict the function outputs that are not used in training. In the second part, apply the Model to predict the output of the Mackey-Glass equation [11]. Next, apply the Model to the real data collected from an electric power plant in the literature. Finally, assess the performance of our Model by comparing it with the results of a similar model implemented in the "*Neuro-Fuzzy Designer*" of MATLAB toolbox.

### A. Model evaluation by the Mackey-Glass equation

In this section, investigate the performance of our ANFIS model by applying it to the chaotic Mackey-Glass differential equation [11]:

$$\dot{x}(t) = \frac{0.2x(t-17)}{1+x^{10}(t-17)} - 0.1x(t) \qquad \text{Eq.4}$$

To obtain the time series value at each integer time point, solving the equation numerically using the fourth-order Runge-Kutta method with the initial condition $x(0) = 1.2$. Therefore, $x(t)$ is obtained for the time range of $0 \leq t \leq 2000$. Then extracted 2000 input-output data pairs in the following format:

$$[x(t-12) \quad x(t-6) \quad x(t); \quad x(t+6)] \qquad \text{Eq. 5}$$

The first 630 data were used to train the ANFIS, and the remaining 370 pairs were used to test the Model. The Model has three inputs, and three membership functions were assigned to each input. The desired and predicted values of both training and checking data are shown in Figure II. As it can be seen, the ANFIS model successfully predicts the time series data extracted from the Mackey-Glass differential equation.

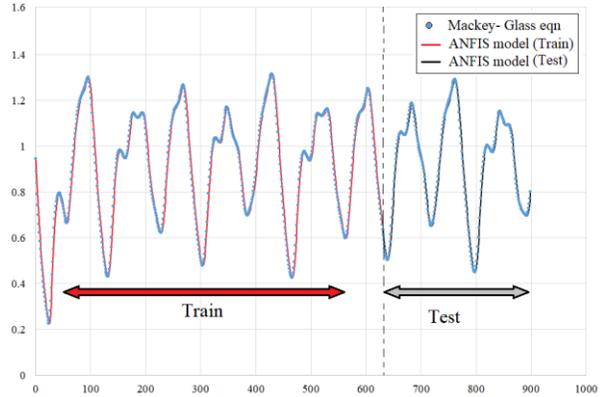

Figure.II Six steps ahead prediction of the Mackey-Glass time series equation by ANFIS model designed in this study.

### B. Model evaluation with actual data

The database used to make the ANFIS models in this study was obtained from the UC Irvine Machine Learning Repository website [21]. The Model is applied to the data available for a combined cycle power plant in Turkey collected in 2015. The 3D CAD design of a combined-cycle power plant is demonstrated in Figure, where the CAD is taken from "www.power-technology.com".

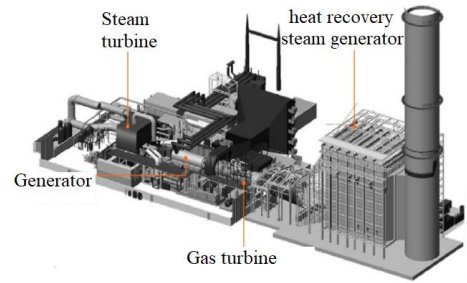

Figure.III -3D CAD design of a combined-cycle power plant taken from www.power-technology.com.

The plant uses gas and steam turbines to increase process efficiency. The disposed heat from the gas turbine is directed to a nearby steam turbine, which can generate extra power. The plant consists of many complicated components. Therefore, obtaining an analytical solution for predicting the plant's power output is very difficult. Here, ANFIS is used to indicate the electricity generation of the plant base on three inputs.

The dataset represents the generated electrical power of the plant, $E$ (MW), as a function of the ambient temperature, $T$ (K), ambient pressure $P$ (mbar), and relative humidity, $H$ (%). As claimed by the data publisher, the data was shuffled five times. For each shuffling, 2-fold cross-validation is carried out to ensure they are randomly distributed. The data were divided into two sets: the training and the checking data set. The graphical summary of the database is presented in Figure.*V*. As it can be seen. At the same time, the output power has a distinguishable relation with the temperature; it has no apparent relation with the pressure and the relative humidity. Next, use ANFIS to approximate the connections between the inputs and the output, even though it is a difficult task for human beings to do so.

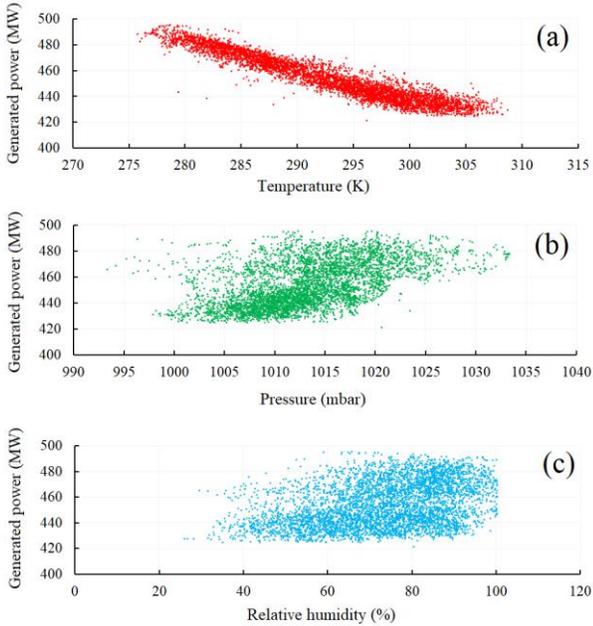

Figure.V Measured data of generated electrical power in MW as a function of (a) temperature, (b) pressure, and (c) relative humidity.

The system was implemented using the procedure explained previously, and was optimized using the hybrid optimization method discussed earlier. Figure illustrates the membership functions before and after training. The dashed curves show the initial membership functions, while the solid curves are the final membership functions. As it can be seen, all membership functions have changed due to training. Here, each variable has three membership functions and, therefore, the linguistic variables of "low" may be assigned to Mf1, "medium" to Mf2, and "high" to Mf3 for each input. The location of the maximum represents the member with the highest contribution. For example, in the temperature plot, 294 K is counted as a medium temperature, and Mf2 can represent temperatures around 294 K. The same is true for the pressure and relative humidity. A pressure of 995 mbar and relative humidity of 28% are both counted as low quantities in a combined cycle power plant. The trained membership functions must have some characteristics. First, they should be continuous. This means that any number in the universe of discourse should be presented by at least one of the membership functions. Gaussian Mfs are guaranteed to satisfy this condition as they span from $-\infty$ to $+\infty$. However, care should be taken not to allow minimal values for the multiplications of the Mfs. Second, there should be no gap between the membership functions. As can be seen in Figure, the final Mfs satisfy this criterion. The Mfs for the second input (pressure) have become slightly squeezed. To improve the ANFIS performance, restrictions may be applied to the variance of all Mfs so that the intersection of all Mfs is always larger than a threshold (e.g. $\mu$intersection $\geq 0.5$). This condition has not been applied in this study. As can be seen in Figure, the membership functions have not undergone a significant change. This is expectable since the number of train data is relatively small. In this case, as Jang [4] mentioned, fixed membership functions could also be used throughout the learning process.

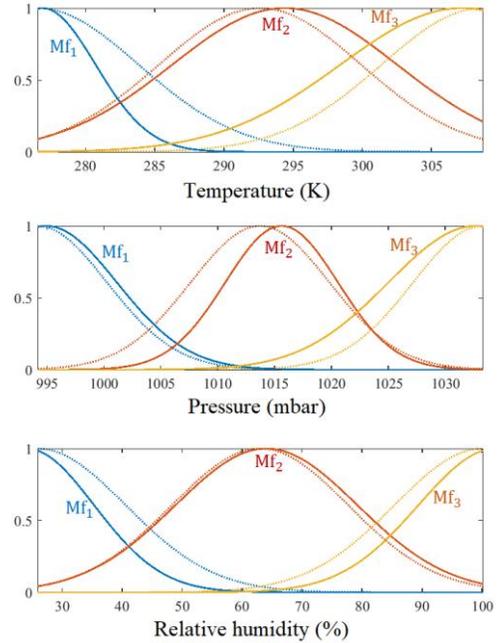

Figure VI. The membership functions before (dashed curves) and after (solid curves) training.

The trained data, test data, as well as model predictions are compared in FigureV. The first and the third panels show the result of the custom code used in our study. The second and fourth panels are the results of the MATLAB ANFIS toolbox. It can be seen that both approaches give almost the same plots, which confirm the validity of the MATLAB code. In each method, the termination criterion was set to be fulfilled when the error was below $10^{-5}$. The number of iterations was not the same for both approaches as the Matlab Toolbox uses its optimized learning rate, which may differ from that set in our code. Interestingly, both RMSEs of the fitted Model to the train and the test data in our code were slightly better than those of the Matlab Toolbox. Besides, the simulation time per epoch in our study (0.16 s per epoch) was shorter than that of the Toolbox (0.55 s per epoch), as summarized in *table-I*.

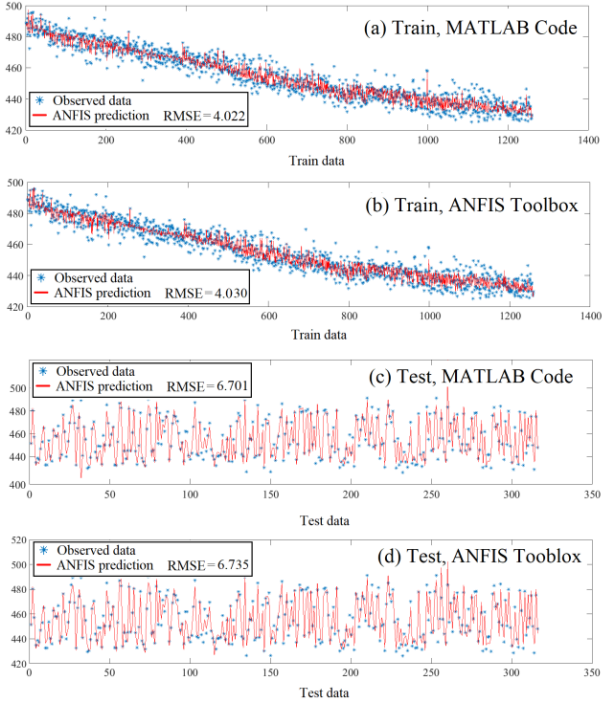

FigureVII Model predictions of the trained data and the tested data.

To get a better picture of the model prediction, the three-dimensional plots of the output against the pairs of inputs are illustrated in figure below. The scattered data shown by the black dots are the actual outputs, and the surface plots are the ANFIS predictions which have been fitted to the data quite well.

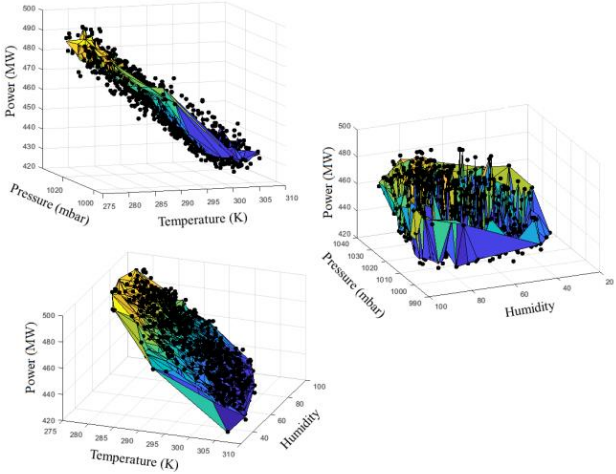

Figure VIII– Three-dimensional representation of the ANFIS prediction. Black dots are the actual outputs, and the surfaces are the ANFIS predictions

Table I Comparison of the RMSE of the custom code and MATLAB ANFIS Toolbox

|  | Training data size | Training RMSE | Checking data size | Checking RMSE | Time per epoch (s) |
|---|---|---|---|---|---|
| MATLAB toolbox | 1259 | 4.0299 | 315 | 6.735 | 0.55 |
| ANFIS code | 1259 | 4.0261 | 315 | 6.701 | 0.16 |

Since there is only one output in the ANFIS network used here, the evaluation of the overall fitness of the output data with the observed ones through a parity plot. Figure shows a parity plot in which the observed data are plotted on the horizontal axis, and the ANFIS predictions are plotted on the vertical axis. A 45° line is also planned to facilitate the comparison. The closer the data are to the 45-degree line, the better the ANFIS prediction is. As the plot shows, most of the data lie at 45°, indicating a close agreement between the targets and the predictions.

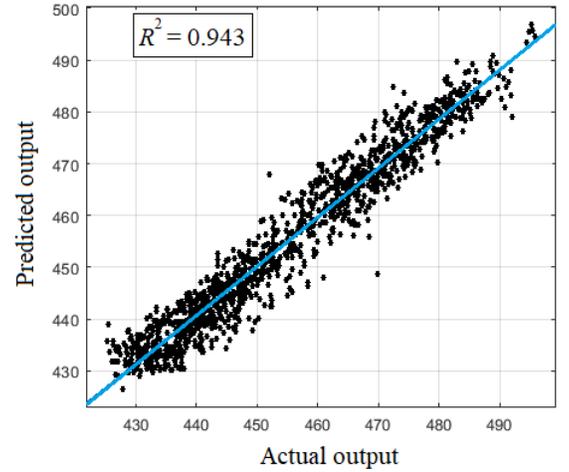

Figure X Scatter plot of the predicted values versus the target values with the corresponding $R^2$ using ANFIS

The correlation coefficient is an index that demonstrates the strength of the relationship between the actual values and the predicted values and is given by:

$$R^2 = \frac{\sum_{k=1}^{P}\left(y_{obs}^{(k)} - \bar{y}_{obs}\right)^2 - \sum_{k=1}^{P}\left(y_{obs}^{(k)} - y_{model}^{(k)}\right)^2}{\sum_{k=1}^{P}\left(y_{obs}^{(k)} - \bar{y}_{obs}\right)^2} \quad \text{Eq. 6}$$

where $\bar{y}_{model}$ and $\bar{y}_{obs}$ are the average values of $y_{model}^{(k)}$ and $y_{obs}^{(k)}$, respectively. The correlation coefficient takes a value between 0 and 1. A model with a higher $R$ is said to have better performance. The correlation coefficient of the plot is 0.943, which is quite close to 1.

## V. Discussion

It has been suggested that the data be normalized or standardized before being introduced into the Model for better convergence. These concepts were applied to the real data.

However, no improvement was seen as a result of normalization and standardization. For some learning rates, the Model diverged. As a result, keep the original form in the training and testing process. This is reasonable since, as shown in Figure.V, the range of data for all inputs does not differ significantly.

## A. Selection of different membership functions

Table-II Different types of membership functions on the error

| Name | MATLAB notation | RMSE (First rder) | RMSE (Zero rder) |
|---|---|---|---|
| Gaussian | gaussmf | 4.096 | 4.374 |
| Triangular | trimf | 4.140 | 4.332 |
| Trapezoidal | trapmf | 4.139 | 6.401 |
| General Bell | gbellmf | 4.084 | 4.486 |
| Pi-shaped | pimf | 4.1537 | 7.723 |
| Difference of 2 sigmoidal | dsigmf | 4.0885 | 4.365 |
| Product of 2 sigmoidal | psigmf | 4.0874 | 4.358 |
| Combination of 2 Gaussian | gauss2mf | 4.0935 | 5.062 |

it can be seen that the Gaussian membership function that was used in this research paper turns out to be the best choice. Trapezoidal and the Pi-shaped membership functions are not good choices for zero-order Sugeno systems since they make poorer predictions than the other types. The triangular membership function would also be a good choice. However, it requires additional programming effort as the sign of the derivatives changes at the function's maximum. As it appears in

Table-II the Finally, the combination of two sigmoidal or two Gaussian membership functions, although they provide a small RMSE, does not significantly improve the error compared to when the individual sigmoidal and Gaussian membership functions are used.

## B. Learning rate ($\eta$)

The learning rate is one of the most critical parameters that should be adjusted in the training process. The learning rate may be initially set to a small value. The learning rate adjusts the magnitude of the weight updates to minimize the loss function. If a small $\eta$ is selected, the Model will probably converge. Still, it takes a lot of time to do so because steps towards the minimum of the loss function are minimal. On the other hand, if the learning rate is too large, training may not converge or even diverge as the solution may jump over the minimum. If the learning rate is allowed to change during the epochs, the system can find the answer much faster.

The training error curves with different learning rates are shown in figure below. At the end of the training with each $\eta$, the error curves are evaluated. If the loss increases, a smaller $\eta$ is selected since it is a sign of the performance degradation, which can lead to overfitting. The overfitting issue will be addressed later. By having a closer look at the error in figure below, it can be seen that the error values for $\eta = 0.003$ and for $\eta = 0.004$ increase after 200 and 300 epochs, respectively. This implies that these learning rates may not be appropriate for this specific dataset and may result in overfitting. However, for $\eta = 0.002$, the error is monotonously decreasing. As a result, $\eta = 0.002$ was selected for the learning rate and unless otherwise stated, this value was used in the rest of the study to analyze the ANFIS performance. 60epochs to satisfy the error condition. However, RMSE is another factor that should be considered, as different learning rates result in different RMSEs. This approach has been proposed by Jang [4]. In this method, the learning rate increases after four successive declines in the error and decreases after two combinations of 1 up and one down. It can be seen that the best RMSE (4.0126) is obtained by applying this method after 118 epochs only. This approach is even better than case 1, where the initial learning rate is seven times larger. The plot of the error with the iteration number for case 9 is shown in figure X . Effect of varying $\eta$ on the performance of the ANFIS code. If the learning rate is allowed to change during the epochs, the system can find the solution much faster. For instance, by setting the initial learning rate to 0.05 and reducing it by 80% in every five epochs (case 6), it only takes

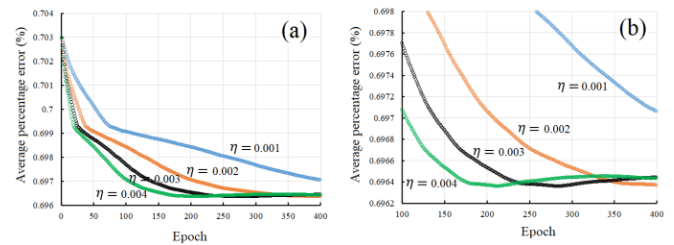

Figure-X Comparison of convergence speed of the NF predictors trained with different learning rates. Training is stopped if the error increases after certain epochs to prevent overfitting.

## C. Overfitting

Overfitting is one of the most common issues in machine learning. It happens when a model is trained too long so that it learns the unnecessary detail and noise in the training data. Since this behaviour may not exist in the new data, it deteriorates the ability of the Model to be generalized to new datasets. One way to minimize the overfitting issue is to check the RMSE of the test data after a certain number of epochs. Once it starts to increase, the training should be stopped. At this point, the Model has the best skill in both the training and unseen test datasets. This method adds computational cost during training which can be reduced by evaluating the Model less frequently, such as every ten training epochs. The RMSE of the train and the test datasets vs the epoch number is shown in figure XI. It can be seen that the train RMSE decreases rapidly up to epoch 300 and continues to decrease gradually after the 300 epochs. However, the RMSE of the test data has a minimum of 650 epochs. This means that after 650 epochs, further iterations worsen the learning process. Therefore, the number of epochs should not exceed 650 in this study to avoid overfitting.

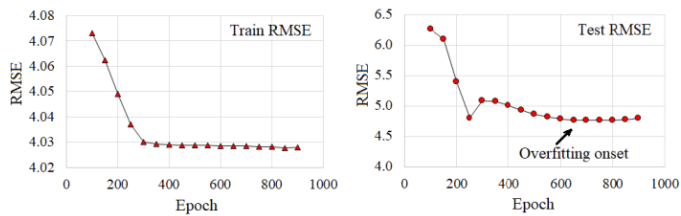

Figure -XI RMSE of the train and the test datasets as a function of the number of epochs.

## VI-Conclusion

This study presented an ANFIS architecture to predict the output electrical power generated by a combined cycle power plant. The system had three inputs: temperature, pressure, and relative humidity. The ANFIS linear and nonlinear parameters were optimized using the hybrid method that uses gradient descent and LSE simultaneously. A first-order type Sugeno model was used to de-fuzzified the output of the aggregated fuzzy set.

The ANFIS model was constructed in MATLAB by developing a custom code. Before predicting the data, the validity of the code was confirmed by applying it to approximate a nonlinear equation with three variables and the ANFIS toolbox in MATLAB. Excellent agreements were found between the code results and those produced by the MATLAB toolbox. The code had a better performance than the MATLAB toolbox in the accuracy of the predictions. Furthermore, the custom code was much faster than the MATLAB toolbox, which may be because it is more straightforward and written for a specific application.

The effect of different parameters of ANFIS on its performance and predictions were studied. Different learning rates were used to simulate the system, and the strategy to select the best learning rate was discussed. Data normalization, usually recommended, was not helpful in our study as it led to divergence on some occasions. It was found that the more significant number of membership functions assigned to each variable can reduce the RMSE of the predicted data. However, this is accompanied by a magnificent increase in the simulation time and allocated memory, two crucial factors determining the number of membership functions. Besides, different membership functions were analyzed, and the Gaussian, triangular, and general Bell functions seemed to be the best choice for this specific application. The overfitting issue was also addressed. It was found that training the system beyond 600 epochs gradually decreases the ability of the Model to make good predictions of the test data. In conclusion, ANFIS has great potential to be applied to different engineering and non-engineering fields. The results obtained here are promising and encouraging enough to imply that ANFIS, along with optimal data selection strategy for training, should be considered one of the most efficient techniques among the existing conventional modelling techniques.

## References

[1]     B. Bektas Ekici and U. T. Aksoy, "Prediction of building energy needs in early stage of design by using ANFIS," *Expert Syst. Appl.*, vol. 38, no. 5, pp. 5352–5358, May 2011, doi: 10.1016/j.eswa.2010.10.021.

[2]     P. MELIN and O. CASTILLO, "Intelligent control of a stepping motor drive using an adaptive neuro?fuzzy inference system," *Inf. Sci. (Ny).*, vol. 170, no. 2–4, pp. 133–151, Feb. 2005, doi: 10.1016/j.ins.2004.02.015.

[3]     P. Melin, J. Soto, O. Castillo, and J. Soria, "A new approach for time series prediction using ensembles of ANFIS models," *Expert Syst. Appl.*, vol. 39, no. 3, pp. 3494–3506, Feb. 2012, doi: 10.1016/j.eswa.2011.09.040.

[4]     J.-S. R. Jang, "ANFIS: adaptive-network-based fuzzy inference system," *IEEE Trans. Syst. Man. Cybern.*, vol. 23, no. 3, pp. 665–685, 1993, doi: 10.1109/21.256541.

[5]     M. Mohandes, S. Rehman, and S. M. Rahman, "Estimation of wind speed profile using adaptive neuro-fuzzy inference system (ANFIS)," *Appl. Energy*, vol. 88, no. 11, pp. 4024–4032, Nov. 2011, doi: 10.1016/j.apenergy.2011.04.015.

[6]	M. Rezakazemi, A. Mosavi, and S. Shirazian, "ANFIS pattern for molecular membranes separation optimization," *J. Mol. Liq.*, vol. 274, pp. 470–476, Jan. 2019, doi: 10.1016/j.molliq.2018.11.017.

[7]	N. Kasabov and D. Filev, "Evolving Intelligent Systems: Methods, Learning, & Applications," in *2006 International Symposium on Evolving Fuzzy Systems*, Sep. 2006, pp. 8–18, doi: 10.1109/ISEFS.2006.251185.

[8]	S. Mitra and Y. Hayashi, "Neuro-fuzzy rule generation: survey in soft computing framework," *IEEE Trans. Neural Networks*, vol. 11, no. 3, pp. 748–768, May 2000, doi: 10.1109/72.846746.

[9]	M. Alizadeh, F. Jolai, M. Aminnayeri, and R. Rada, "Comparison of different input selection algorithms in neuro-fuzzy modeling," *Expert Syst. Appl.*, vol. 39, no. 1, pp. 1536–1544, Jan. 2012, doi: 10.1016/j.eswa.2011.08.049.

[10]	S. Barak, J. H. Dahooie, and T. Tichý, "Wrapper ANFIS-ICA method to do stock market timing and feature selection on the basis of Japanese Candlestick," *Expert Syst. Appl.*, vol. 42, no. 23, pp. 9221–9235, Dec. 2015, doi: 10.1016/j.eswa.2015.08.010.

[11]	M. Mackey and L. Glass, "Oscillation and chaos in physiological control systems," *Science (80-. ).*, vol. 197, no. 4300, pp. 287–289, Jul. 1977, doi: 10.1126/science.267326.

[12]	S. Mitaim and B. Kosko, "What is the best shape for a fuzzy set in function approximation?," in *Proceedings of IEEE 5th International Fuzzy Systems*, vol. 2, pp. 1237–1243, doi: 10.1109/FUZZY.1996.552354.

[13]	Jin Zhao and B. K. Bose, "Evaluation of membership functions for fuzzy logic controlled induction motor drive," in *IEEE 2002 28th Annual Conference of the Industrial Electronics Society. IECON 02*, vol. 1, pp. 229–234, doi: 10.1109/IECON.2002.1187512.

[14]	A. Al-Hmouz, Jun Shen, R. Al-Hmouz, and Jun Yan, "Modeling and Simulation of an Adaptive Neuro-Fuzzy Inference System (ANFIS) for Mobile Learning," *IEEE Trans. Learn. Technol.*, vol. 5, no. 3, pp. 226–237, Jul. 2012, doi: 10.1109/TLT.2011.36.

[15]	A. Sadollah, "Introductory Chapter: Which Membership Function is Appropriate in Fuzzy System?," in *Fuzzy Logic Based in Optimization Methods and Control Systems and its Applications*, InTech, 2018.

[16]	M. Sugeno and G. . Kang, "Structure identification of fuzzy model," *Fuzzy Sets Syst.*, vol. 28, no. 1, pp. 15–33, Oct. 1988, doi: 10.1016/0165-0114(88)90113-3.

[17]	W. Zeng, Y. Zhao, and Q. Yin, "Sugeno fuzzy inference algorithm and its application in epicentral intensity prediction," *Appl. Math. Model.*, vol. 40, no. 13–14, pp. 6501–6508, Jul. 2016, doi: 10.1016/j.apm.2016.01.065.

[18]	Soteris A. Kalogirou, *Solar Energy Engineering*. Elsevier, 2009.

[19]	M. Panella and A. S. Gallo, "An input-output clustering approach to the synthesis of ANFIS networks," *IEEE Trans. Fuzzy Syst.*, vol. 13, no. 1, pp. 69–81, Feb. 2005, doi: 10.1109/TFUZZ.2004.839659.

[20]	H. Takagi and I. Hayashi, "NN-driven fuzzy reasoning," *Int. J. Approx. Reason.*, vol. 5, no. 3, pp. 191–212, May 1991, doi: 10.1016/0888-613X(91)90008-A.

[21]	"UC Irvine Machine Learning Repository." https://archive.ics.uci.edu/ml/datasets/Combined+Cycle+Power+Plant.